\documentclass[preprint,12pt]{elsarticle}

\usepackage{amssymb}
\usepackage{amsmath}
\usepackage{multirow}
\usepackage{booktabs}
\usepackage{color}
\usepackage{balance}
\usepackage{bm}
\usepackage{subfigure}
\usepackage[table, xcdraw]{xcolor}
\usepackage{makecell}

\journal{NeuroComputing}

\begin{document}

\begin{frontmatter}



\title{Neural Augmentation Based Panoramic High Dynamic Range Stitching} %% Article title

\author[1]{Chaobing Zheng }
\ead{zhengchaobing@wust.edu.cn}

\author[2]{Yilun Xu}
\ead{yilunxu\_buaa@163.com}

\author[2]{Weihai~Chen}
\ead{whchen@buaa.edu.cn}

\author[1]{Shiqian Wu}
\ead{shiqian.wu@wust.edu.cn}

\author[3,4,5]{Sen Zhang}
\ead{zhangsen@ustb.edu.cn}

\author[6]{and Zhengguo Li}
\ead{ezgli@i2r.a-star.edu.sg}

\address[1]{Institute of Robotics and Intelligent Systems, School of Electronic Information, Wuhan University of Science and Technology, 947 Heping Avenue, Wuhan 430081, China}

\address[2]{School of Automation Science and Electrical Engineering, Beihang University, Beijing, 100191, China}

\address[3]{School of Automation and Electrical Engineering, University of Science and Technology Beiing, Beijing 100083, China}

\address[4]{Key Laboratory of Knowledge Auomalion for indusrial Proceses of Mimistry ofEducation, University of Science and Techmology Beijing, Beijimg 1083, China}

\address[5]{Shunde lnnovation School, University of Science and Technology Beijing, Foshan 528300, China}

\address[6]{VI Department, Institute for Infocomm Research, A*STAR, 138632, Singapore}

%% Abstract
\begin{abstract}
Due to saturated regions of inputting low dynamic range (LDR) images and large intensity changes among the LDR images caused by different exposures, it is challenging to produce an information enriched panoramic LDR image without visual artifacts for a high dynamic range (HDR) scene through stitching multiple geometrically synchronized LDR images with different exposures and pairwise overlapping fields of views (OFOVs).  Fortunately,  the stitching of  such images is innately a perfect scenario for the fusion of a physics-driven approach  and a data-driven approach  due to their OFOVs. Based on this new insight, a novel neural augmentation based panoramic HDR stitching algorithm is proposed in this paper. The physics-driven approach is built up using the OFOVs. Different exposed images of each view are initially generated by using the physics-driven approach, are then refined by a data-driven approach, and are finally used to produce panoramic LDR images with different exposures. All the panoramic LDR images with different exposures are combined together via a multi-scale exposure fusion algorithm to produce the final panoramic LDR image. Experimental results demonstrate the proposed algorithm outperforms existing panoramic stitching algorithms.
\end{abstract}

\begin{highlights}
\item Research highlight 1:
A new R\&D problem is introduced for generating a set of panoramic LDR images with different exposures from the same number of geometrically synchronized LDR images with different exposures and OFOVs. The problem is addressed by a novel neural augmentation based framework which contains an IMF-based method and an CNN-based approach. Our neural augmentation framework converges faster than the data-driven approach. The framework can be applied to develop a new type of panoramic stitching systems;

\item Research highlight 2:
An elegant histogram bins-based IMF estimation algorithm which is more accurate than existing IMF estimation algorithms and is more robust than the algorithms with respect to camera movements and moving objects;

\item Research highlight 3:
A simple physics-driven approach to produce a panoramic LDR image with enriched information from a set of geometrically aligned and differently exposed LDR images with OFOVs. The genuine high-frequency information is effectively preserved in the final image.

\end{highlights}

%% Keywords
\begin{keyword}


High dynamic range imaging, panoramic stitching, computational photography, weighted histogram averaging, neural augmentation,  multi-scale exposure fusion
\end{keyword}

\end{frontmatter}


\section{Introduction}
Panoramic stitching can compose a group of images with  pairwise overlapping fields of views (OFOVs) from a real-world scene to synthesize an image with a wider view for the scene \cite{chen2024hdr}. All these images are usually 8-bit sRGB images and their exposure times are almost the same. Such a capturing method performs well for those real-word low dynamic range (LDR) scenes. However, a real-world high dynamic range (HDR) scene could have a dynamic range of 100,000,000:1 whereas an 8-bit sRGB image only has a dynamic range of 256:1 \cite{CRFs,1aggarwal2001, li2024generalizing, yan2021towards, yan2024}. Our human eyes are able to  perceive dynamic ranges by up to 1,000,000:1. There are underexposed and/or overexposed regions in panoramic LDR images which are synthesized via the existing panoramic LDR stitching. Therefore, the inputting images of panoramic stitching are highly demanded to be captured by using different exposures for those real-world HDR scenes to give users a higher quality of experience \cite{1aggarwal2001,1Eden2006,Yan2024, Li2024, Xu2022}. The resultant panoramic stitching is a new type, and can be called panoramic HDR stitching.

\begin{figure*}[htb]
	\setlength{\abovecaptionskip}{0pt}
	\setlength{\belowcaptionskip}{0pt}
	\begin{center}
		\includegraphics[width=1\linewidth]{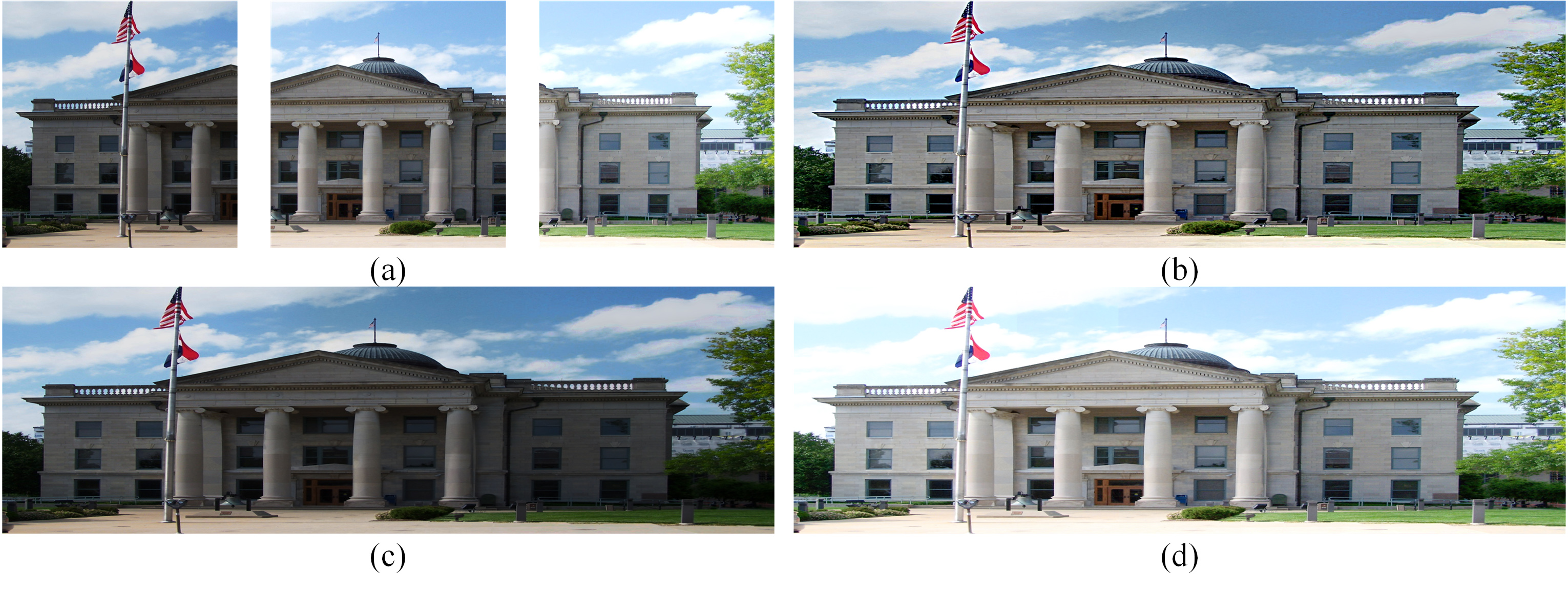}
	\end{center}
	\caption{Illustration of panoramic LDR and HDR imaging. (a) are three differently exposed LDR images with OFOVs and exposure value (EV) gaps as 1's; (b) is an output image by the proposed HDR stitching algorithm; as well as (c) and (d) are two output images by using an existing LDR panoramic imaging algorithm.}
	\label{necessarityexample}
\end{figure*}

Two main steps of panoramic HDR stitching are: 1) to geometrically register all the inputting images by extending those algorithms in \cite{nie2023parallax, Zhao2023, Zhao2022}, and 2) to stitch all the aligned images together photometrically to removal all possible visual artifacts and recover underexposed and overexposed information. \textcolor{blue}{This paper primarily concentrates on the latter,} which presents a new problem which has three main challenges: 1) eliminate brightness discrepancies among all the aligned images in the OFOVs caused by the varying exposures; 2) restore lost information in saturated regions of all the aligned images due to very limited number of different exposures, especially in those non-OFOV regions with only one exposure; and 3) preserve fine details of the darkest and brightest areas well in the final panoramic LDR image. These challenges cannot be adequately solved via existing panoramic LDR stitching algorithms \cite{Brown,1ding2021,Jia, wu2022litmnet}. For instance, if all the synchronized images are tuned according to the brightness of the image with the largest (smallest) exposure, the brightest (darkest) areas of the HDR scene are then become overexposed (underexposed) in the panoramic LDR image, as illustrated in Figs. \ref{necessarityexample}(c) and \ref{necessarityexample}(d). Therefore, there is a need to introduce a different panoramic stitching algorithm tailored for real-world HDR scenes.

On top of our conference paper \cite{10312264}, a new type of panoramic stitching algorithm is proposed in this paper for a set of geometrically aligned LDR images with differing exposures and OFOVs, aimed at creating a panoramic LDR image with enriched information in the darkest and brightest regions for an HDR scene. The proposed  panoramic HDR stitching algorithm is built upon a distinctive characteristic of the geometrically aligned LDR images with OFOVs and different exposures: all the pixels in the non-OFOV regions have only one exposure, whereas all the pixels in the OFOVs have two different exposures. Multiple panoramic LDR images with different exposures are generated from the inputs by using a novel physics-driven deep learning framework called neural augmentation. The framework is a seamless integration of a physics-driven approach  and a data-driven approach . The former is on top of intensity mapping functions (IMFs) and the latter is on top of convolutional neural networks (CNNs). The proposed framework effectively combines both the generalization capability of the former and the learning capability of the latter.

The IMFs are estimated for adjacent images using the OFOVs among them, rather than being computed from camera response functions (CRFs) as in previous works \cite{1zheng2020, CMF-PCRF2018}. In other words, the proposed neural augmentation does not assume  the availability of CRFs as in \cite{1zheng2020}. Therefore, the stitching of these images is a good case for the combination of the physics-driven approach and data-driven approach due to the OFOVs. The IMFs are estimated by using a new algorithm named as weighted histogram averaging (WHA) in \cite{1xuy2022} which is based on a novel correspondence between the two images at histogram-bin-level: each histogram bin from one image matches to one unique segment of the histogram bins from the other image. The mapped value is computed by averaging intensities in the matched segment with properly defined weights. The WHA outperforms those methods in \cite{1ding2021, CMF-PCRF2018, CMF-TG2014, CMF-AM2014, CMF-3MS2013, CMF-GPS2017, CMF-2001, Zhu, Grossberg2003}. The IMFs are adopted to initially produce the panoramic LDR images with different exposures. Such a method is a typical physics-driven approach and experimental results indicate that the accuracy of IMFs plays an important role in our neural augmentation. The OFOVs are thus required to be sufficiently large so as to estimate the accurate IMFs. The physics-driven approach is independent of training data, and it usually increases the robustness of our neural augmentation framework if it is accurate.

Even though the physics-driven approach is independent of training data, differently exposed images of each view generated by the  physics-driven approach should be enhanced because of the limited representation capability of the  physics-driven approach. A data-driven approach is thus introduced to refine these images by learning the remaining (or residual) information by proposing a new multi-scale exposedness aware network (MEAN). Our MEAN is designed by incorporating an exposedness aware guidance branch (EAGB) into short-skip connection based recursive residual groups (SRRGs). As pointed out in \cite{1raha2018}, deep CNNs generally have limited capability to represent high-frequency information. The function of both the multi-scale feature refinement structure and the short-skip connection structure is to preserve the desired high-frequency information. The exposedness of the input images is utilized to define a set of binary masks. The EAGB is designed on top of the binary masks, and it can assistant  the proposed MEAN restore the overexposed and underexposed regions more efficiently.

All the differently exposed images of all the views are used to construct panoramic LDR images with different exposures. The generated panoramic LDR images with different exposures are ultimately combined together via the multi-scale exposure fusion (MEF) algorithm in \cite{1kou2017}, together with a  physics-driven approach for the enhancement of high-frequency information. \textcolor{blue}{ It should be pointed out that the MEF algorithm in  \cite{Ma2022} can also be applied here.} This process results in a panoramic LDR image with enriched information. As shown in Fig. \ref{necessarityexample}(b), information in the darkest and brightest regions is preserved better in the stitched image by the proposed algorithm. The proposed algorithm is different from the algorithms in \cite{1aggarwal2001,1Eden2006} in the sense that a panoramic HDR image is first generated and then being tone mapped to a panoramic LDR image for displaying by the algorithms in \cite{1aggarwal2001,1Eden2006}. It is also different from the algorithms in \cite{Brown,1ding2021,Jia} in the sense that the algorithms in \cite{Brown,1ding2021,Jia} focus on real-world LDR scenes. Overall, the major contributions of this paper can be summarized as follows:

1) a new R\&D problem is introduced for generating a set of panoramic LDR images with different exposures from the same number of geometrically synchronized LDR images with different exposures and OFOVs. The problem is addressed by a novel neural augmentation based framework which contains an IMF-based method and an CNN-based approach. Our neural augmentation framework converges faster than the data-driven approach. The framework can be applied to develop a new type of panoramic stitching systems;

2) an elegant histogram bins-based IMF estimation algorithm which is more accurate than existing IMF estimation algorithms and is more robust than the algorithms with respect to camera movements and moving objects;

3) a simple physics-driven approach to produce a panoramic LDR image with enriched information from a set of geometrically aligned and differently exposed LDR images with OFOVs. The genuine high-frequency information is effectively preserved in the final image.

\section{An Panoramic HDR Stitching Dataset}
\label{dataset}
Same as other data driven research topics, data set plays an important role in the panoramic HDR stitching. Our panoramic HDR stitching data-set is built up from three well known HDR imaging data sets in \cite{1zheng2020,1cai2018,1xuyl2021}. All these differently exposed images in these datasets are captured in real-world. Each group of LDR images is constructed to resemble captures from a location with OFOVs, representing an HDR scene via a wide-angle perspective. \textcolor{blue}{Images are captured by two different cameras which are VETHDR-Nikon and VETHDR-Canon, exposure times are changed, the interval between the two adjacently exposed images is 1-EV. And there is a certain overlap between adjacent images.}

\begin{figure*}[htb]
	\centering
	\includegraphics[width=1\textwidth]{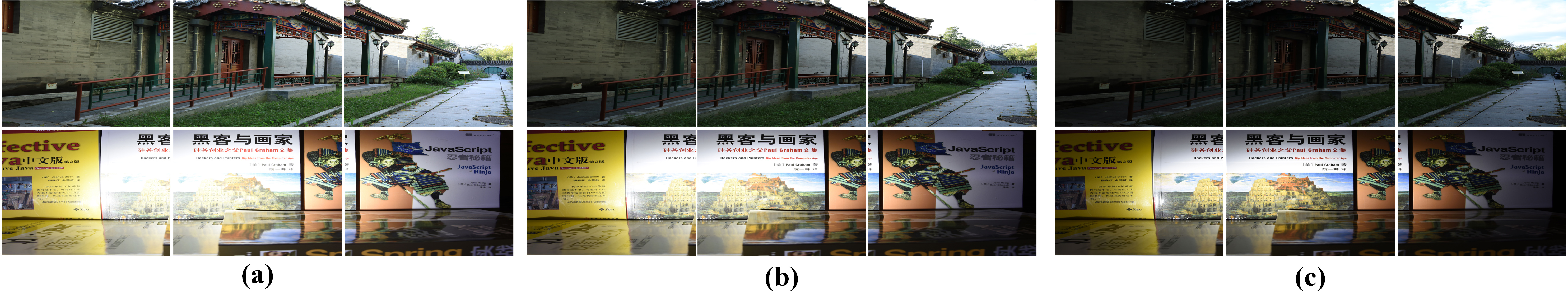}
	\caption{Two examples in the proposed panoramic HDR stitching dataset. (a) are images with high exposures. (b) are images with middle exposures. (c) are images with low exposures.}
	\label{data}
\end{figure*}

Nine images are captured for each HDR scene by using three different exposures and three different orientations. Three sets of LDR images with different exposures are captured for each orientation of the HDR scene. The nine images are denoted as ($Z_1$, $Z_T^{1 \to 2}$,  $Z_T^{1 \to 3}$), ($Z_T^{2 \to 1}$, $Z_2$, $Z_T^{2 \to 3}$), and ($Z_T^{3 \to 1}$, $Z_T^{3 \to 2}$, $Z_3$). Neither  moving objects nor camera movements exist in the three LDR images  with different exposure for each orientation. Our data set consists of 780 training sets, 80 validation sets, and 120 test sets. Two sets are illustrated in Fig. \ref{data}.

Inputs of the proposed panoramic HDR stitching algorithm are three differently exposed LDR images with OFOVs. They are $Z_1$, $Z_2$, and $Z_3$. $\chi_i(1\leq i\leq 3)$ is the non-OFOV region which is viewed only in the image $Z_i(1\leq i\leq 3)$. The corresponding sub-image is denoted as $Z_i^{no}$.  The two pairwise OFOVs are denoted as $\Xi_1^2$ and $\Xi_2^3$. The corresponding sub-images are $Z_{1,2}^o$, $Z_{2,1}^o$, $Z_{2,3}^o$ and $Z_{3,2}^o$. It can be easily verified that the panoramic HDR image covers the spatial area of $(\chi_1\cup \Xi_1^2\cup \chi_2\cup \Xi_2^3\cup \chi_3)$, and the images $Z_1$, $Z_2$ and $Z_3$ can be decomposed as
\begin{align}
	Z_1&=Z_1^{no}\cup Z_{1,2}^o;\\
	Z_2&=Z_{2,1}^o\cup Z_2^{no}\cup Z_{2,3}^o,\\
	Z_3&=Z_{3,2}^o\cup Z_3^{no}.
\end{align}

Each LDR image has a resolution of $480\times 640$, and the overlapping area has a resolution of $480\times 200$. It should be pointed out that the resolution of the overlapping area is important for the proposed neural augmentation framework. The exposure times are $\Delta t_l(1\leq l\leq 3)$ which satisfy $\Delta t_3=2\Delta t_2=4\Delta t_1$. In other words, the EV gap between the image $Z_i$ and $Z_{i+1}$ is 1.

The inputs of the proposed panoramic HDR stitching possess a distinctive property: all the pixels in the non-OFOVs only have one exposure, while all the pixels in the OFOVs have two different exposures. The inputs are different from the input in singe image based HDR imaging \cite{1bant2006,DrTMO,1zheng2020} where all pixels only have one exposure. This is a new problem for the data-driven approach. It is difficult to restore three panoramic LDR images with different exposures from the three images. A new neural augmentation framework will be introduced to address this challenging problem in the next section.

\section{Neural Augmented Generation of Panoramic LDR Images with Different Exposures}
\label{newalgorithm}

\begin{figure*}[htb]
	\centering
	\includegraphics[width=1.0\textwidth]{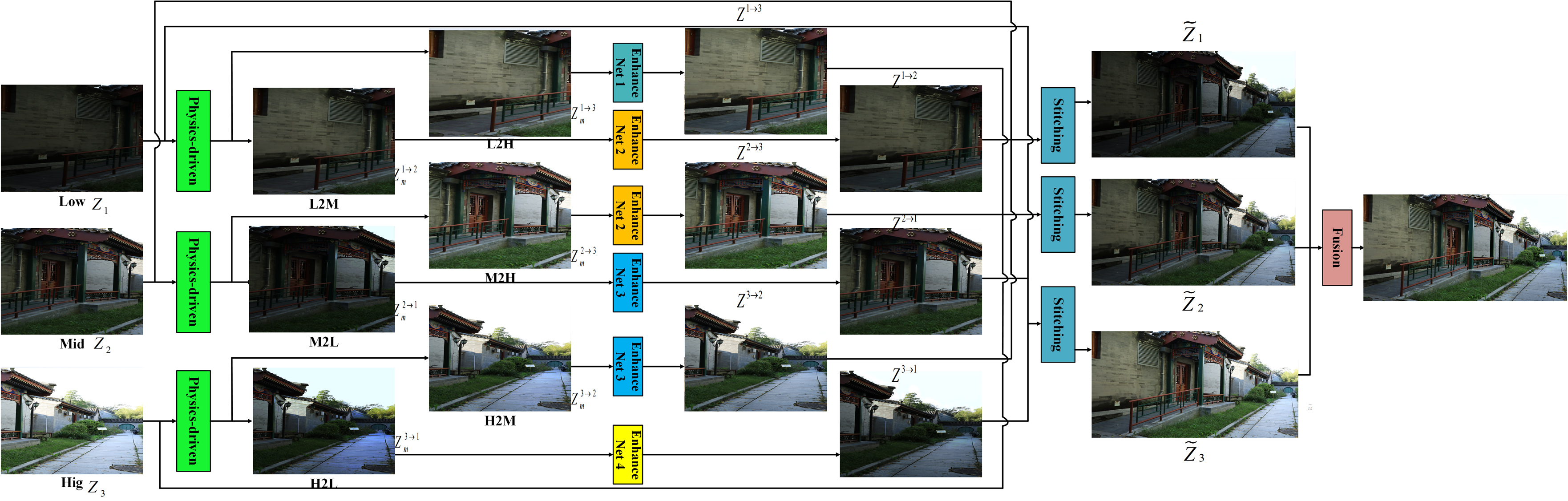}
	\caption{The diagram of our panoramic HDR imaging. Three panoramic LDR images with different exposures are generated from three differently exposed LDR images with different orientations via a physics-driven deep learning framework. They are combined together to generate a panoramic LDR image with enriched information.}
	\label{Fig_1}
\end{figure*}

Diagram of our panoramic HDR imaging is shown in Fig. \ref{Fig_1}. Three differently exposed panoramic LDR images are synthesized via a novel neural augmentation framework which seamlessly integrates physics-driven and data-driven approaches. They will be combined together to generate an information enriched panoramic LDR image.

Two more differently exposed images are generated for image $Z_i$  $(1\leq i\leq 3)$  through
\begin{align}
	\nonumber
	Z^{i\to j}=&f^{i\to j}(Z_i)\\ \label{brightnesseq1}
	=&f_m^{i\to j}(Z_i)+f_{d}^{i\to j}(Z_i), 1\leq i\neq j\leq 3,
\end{align}
where $f_m^{i\to j}(\cdot)$ is a physics-driven approach, and $f_d^{i\to j}(\cdot)$ is a data-driven approach.

The three panoramic LDR images $\tilde{Z}_l(l\in\{1,2,3\})$ are then generated from the images $Z_1$, $Z_2$, $Z_3$ and $Z^{i\to j}$'s $(1\leq i\neq j\leq 3)$.  When the first panoramic LDR image $\tilde{Z}_1$ is  generated,  the brightness of $Z_2$ and $Z_3$ is mapped to that of $Z_1$. The first image  $\tilde{Z}_1$ is produced as
\begin{align}
	\label{fusioneq1}
	\left\{\begin{array}{ll}
		Z_1(p); &\mbox{if~} p\in \chi_1\\
		\Phi_{2,1}(p)Z_1(p)+(1-\Phi_{2,1}(p))Z^{2\to 1}(p); &\mbox{if~}p\in \Xi_1^2\\
		Z^{2\to 1}(p);& \mbox{if~}p\in \chi_2\\
		\Phi_{3,2}(p)Z^{2\to 1}(p)+(1-\Phi_{3,2}(p))Z^{3 \to 1}(p); &\mbox{if~}p\in \Xi_2^3\\
		Z^{3\to 1}(p); & \mbox{otherwise}\\
	\end{array}
	\right.,
\end{align}
where $\Phi_{2,1}(p)$ is defined according to the pixel $p$ in the set $\Xi_1^2$. It approaches 0 when the pixel $p$ moves to the region $\chi_2$, and 1 when $p$ moves to the region $\chi_1$. $\Phi_{3,2}(p)$ is defined according to the pixel $p$ in the set $\Xi_2^3$. It approaches 0 when $p$ moves to the region $\chi_3$, and 1 when the pixel $p$ moves to the region $\chi_2$.

Similarly, the second image  $\tilde{Z}_2$ is generated as
\begin{align}
	\label{fusioneq0}
	\left\{\begin{array}{ll}
		Z^{1 \to 2}(p); &\mbox{if~} p\in \chi_1\\
		\Phi_{2,1}(p)Z^{1 \to 2}(p)+(1-\Phi_{2,1}(p))Z_2(p); &\mbox{if~}p\in \Xi_1^2\\
		Z_2(p);& \mbox{if~}p\in \chi_2\\
		\Phi_{3,2}(p)Z_2(p)+(1-\Phi_{3,2}(p))Z^{3 \to 2}(p); &\mbox{if~}p\in \Xi_2^3\\
		Z^{3\to 2}(p); & \mbox{otherwise}\\
	\end{array}
	\right.,
\end{align}
and the third one is synthesized as
\begin{align}
	\label{fusioneq2}
	\left\{\begin{array}{ll}
		Z^{1 \to 3}(p); &\mbox{if~} p\in \chi_1\\
		\Phi_{2,1}(p)Z^{1 \to 3}(p)+(1-\Phi_{2,1}(p))Z^{2\to 3}(p); &\mbox{if~}p\in \Xi_1^2\\
		Z^{2\to 3}(p);& \mbox{if~}p\in \chi_2\\
		\Phi_{3,2}(p)Z^{2\to 3}(p)+(1-\Phi_{3,2}(p))Z_3(p); &\mbox{if~}p\in \Xi_2^3\\
		Z_3(p); & \mbox{otherwise}\\
	\end{array}
	\right..
\end{align}

\begin{figure*}[htb]
	\centering
	\includegraphics[width=1.0\textwidth]{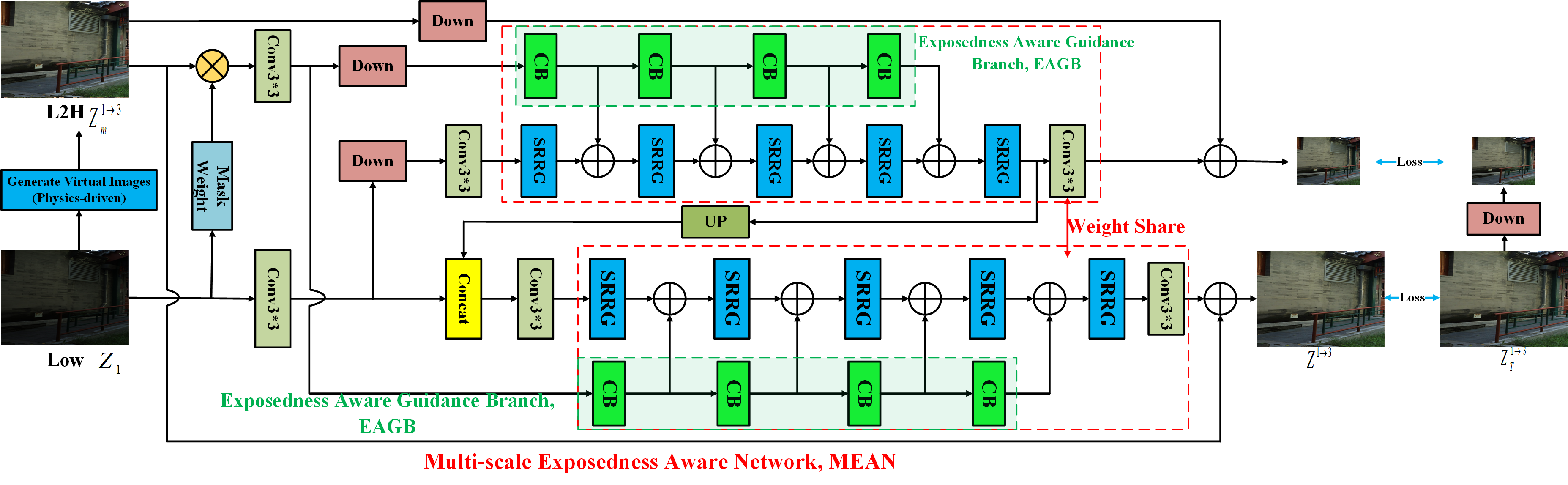}
	\caption{Generation of $Z^{1\to 3}$ (i.e., $i=1$, $j=3$) via the proposed neural augmentation framework. $Z_m^{1\to 3}$ is produced  by using a physics-driven approach and is then refined via a data-driven approach to obtain $Z^{1\to 3}$. Our MEAN is composed of EAGB and SRRG. Our MEAN is on top of the CNN in \cite{CycleISP}. The multi-scale structure can preserve the high-frequency information. The EAGB can help restore the saturated areas more effectively.}
	\label{Fig_3}
\end{figure*}

Details on the $f_m^{i\to j}(Z_i)$ and $f_d^{i\to j}(Z_i)$  for all $i$'s and $j$'s $(1\leq i\neq j\leq 3)$ are given in the following two subsections.

\begin{figure*}[htb]
	\centering
	\includegraphics[width=1.0\textwidth]{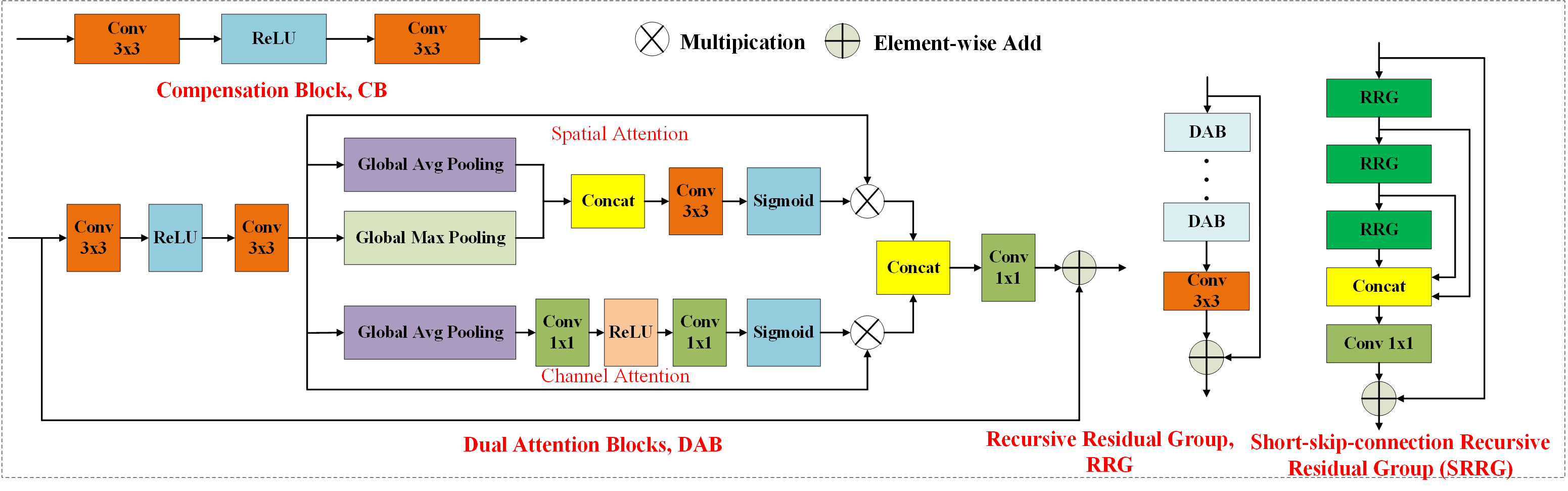}
	\caption{Diagram of the proposed SRRG. Our SRRG is obtained by adding short-skip connections to the RRG in \cite{CycleISP}.}
	\label{Fig_4}
\end{figure*}

\subsection{Physics-driven Initialization}

\label{modelbasedalgo}
The accuracy of the physics-driven approach is crucial for the neural augmentation. Therefore, a novel histogram bins-based algorithm is proposed to estimate the IMFs \cite{1xuy2022}. The physics-driven generation of panoramic LDR images with different exposures builds upon the new IMF estimation algorithm.

The IMFs of the images $Z_i$ and $Z_j$ are computed by using their OFOV. The proposed method is different from the algorithms in \cite{1zheng2020} in the sense that the CRFs are required to be known in advance and the IMFs are computed from the CRFs in \cite{1zheng2020}. Therefore, the generation of panoramic LDR images with different exposures from the differently exposed LDR images with OFOVs is naturally a good case for the application of neural augmentation \cite{1zheng2020}. With some abuse of notation, their OFOVs are also denoted as $Z_i$ and $Z_j$.

Let $p$ be a pixel position.  $\Omega_i(z)$ is defined as $\{p|Z_i(p)=z\}$, and $|\Omega_i(z)|$$(\doteq H_i(z))$ is the number of pixels in  the set $\Omega_i(z)$.  The cumulative histogram $C_i(z)$ of the image $Z_i$ is then computed as $\sum_{k=0}^{z} H_i(k)$. A function $\psi^{i \to j}(z)$ can be defined from the non-decreasing property of the IMFs as \cite{1xuy2022}
\begin{align}
	\label{CH:interval}
	C_j(\psi^{i \to j}(z)-1)< C_i(z) \leq C_j(\psi^{i \to j}(z)),
\end{align}
where $\psi^{i\to j}(-1)$ is  0. With the function $\psi^{i\to j}(z)$, the correspondence between the histogram bins of the images $Z_i$ and $Z_j$ is first built up as follows: the first  and last bins in the image $Z_j$ corresponding to the bin $\Omega_i(z)$ are $\Omega_j(\psi^{i \to j}(z-1))$ and $\Omega_j(\psi^{i \to j}(z))$, respectively.  The cardinalities of sub-bins (or bins) in the image $Z_j$ corresponding to the bin $\Omega_i(z)$ is  denoted by $\hat{H}^{i \to j}(k,z)$. $\hat{H}^{i \to j}(k,z)$ is defined as $H_i(z)$ if $\psi^{i \to j}(z-1)$ is equal to $\psi^{i \to j}(z)$. Otherwise, it is defined as
\begin{align}
	\left\{\begin{array}{ll}
		C_j(k) - C_i(z-1),  &\mbox{if~} k=\psi^{i \to j}(z-1)\\
		C_i(z)-C_j(k-1),  &\mbox{if~}k=\psi^{i \to j}(z)\\
		H_j(k),  &\mbox{otherwise}
	\end{array}
	\right..
\end{align}

The new IMF $\Lambda^{i \to j}(z)$ is then computed by using the correspondance at the  histogram-bin level as	
\begin{align}
	\label{eq:CHCH}
	\Lambda^{i \to j}(z)= \sum_{k=\psi^{i \to j}(z-1)}^{\psi^{i \to j}(z)}\frac{\hat{H}^{i \to j}(k,z)}{H_i(z)}k,  \ H_i(z) \neq0.
\end{align}

Obviously, our new IMF $\Lambda^{i \to j}(z)$ is a weighted average of all the sub-bins (or bins) in the image $Z_j$ corresponding to the bin $H_i(z)$. Therefore, our new IMF estimation algorithm is named as  the WHA. Unlike the pixel-level correspondence in \cite{Zhu},  the correspondence at the histogram-bin-level is robust to moving objects and camera movements. On the other hand, the high-frequency information will be lost when the IMFs are used to synthesize images.

The IMFs $\Lambda^{1 \to 2}(z)$ and $\Lambda^{2 \to 1}(z)$ are computed from the sub-images $Z_{1,2}^o$ and $Z_{2,1}^o$. Similarly, the IMFs $\Lambda^{2 \to 3}(z)$  and $\Lambda^{3 \to 2}(z)$ are obtained from the sub-images $Z_{2,3}^o$ and $Z_{3,2}^o$.  Subsequently, the IMFs  $\Lambda^{1 \to 3}(z)$  and $\Lambda^{3 \to 1}(z)$ can be computed. As shown in the section \ref{IMFaccuracy}, the accuracy of the IMFs plays an important role in the proposed neural augmentation framework. Thus, the resolutions of the sub-images $Z_{1,2}^o$, $Z_{2,1}^o$, $Z_{2,3}^o$ and $Z_{3,2}^o$ are required to be sufficiently large such that  the IMFs $\Lambda^{1 \to 2}(z)$, $\Lambda^{2 \to 1}(z)$, $\Lambda^{2 \to 3}(z)$  and $\Lambda^{3 \to 2}(z)$ are highly accurate. Differently exposed images of each view are generated  as
\begin{align}
	Z_m^{i \to j} = {\Lambda}^{i \to j}(Z_i)\dot{=}f_m^{i\to j}(Z_i), 1\leq i\neq j\leq 3.
\end{align}

Even though the proposed physics-driven approach is independent of training data, its representation capability is usually a little limited. As a result, the quality of images $Z_m^{i \to j}$ $(1\leq i\neq j\leq 3)$  is poor and all the images need to be refined  via a data-driven approach as in the next subsection.

\subsection{Data-driven Refinement}
\label{datadrivenrefinement}
The IMFs are physical models on the correlation among different exposed images. Since their representation capabilities are usually limited, there are unmodelled information. The unmodelled information  $(Z_T^{i \to j} - Z_m^{i \to j})(1\leq i\neq j\leq 3)$  can be further represented by the data-driven approach $f_d^{i\to j}(\cdot)$.

The unmodelled information can be  easily compensated by the $f_d^{i\to j}(\cdot)$ because it is sparser. Since the DNNs cannot learn the high-frequency information well \cite{1raha2018}, it is crucial for the data-driven approach to preserve the high-frequency information. A new MEAN is proposed in Figs. \ref{Fig_3} and \ref{Fig_4} to achieve these objectives. The key components of the proposed MEAN are EAGB, SRRGs, and multi-scale feature refinement. The SRRG  is designed by adding short-skip connections to the RRG in \cite{CycleISP} in order to pass shallow layers' information into deep layers. As such, the high-frequency information has a potential to be well preserved. The proposed SRRG can propagate more informative features among different layers. The multi-scale feature refinement structure can further help preserve the high-frequency information.

The proposed EAGB consists of several compensations blocks (CBs) with each of them including two $3\times 3$ convolutional layers and one activation layer. The EAGB guides the SRRG to restore the saturated areas more effectively by properly binary masks $M_l^{i\to j}$$(1\leq i\neq j\leq 3)$. The masks $M_l^{1\to i}$$(i=2,3) $ are defined for the image $Z_{1,l}$ as
\begin{align}
	\label{M0}
	&M_l^{1\to i}(p) = \left\{ \begin{array}{ll}
		0;&\mbox{if~} Z_{1,l}(p) \leq 5\\
		1;&\mbox{otherwise}
	\end{array} \right..
\end{align}

It should be pointed out that when bright images are mapped to dark images, their binary masks are different. For example, the masks  $M_l^{i\to 1}$$(i=2,3)$  are
\begin{align}
	\label{M2}
	&M_l^{i\to 1}(p) = \left\{\begin{array}{ll}
		0;&\mbox{if~} Z_{i,l}(p) \geq 250\\
		1;&\mbox{otherwise}
	\end{array} \right..
\end{align}

Obviously, the proposed masks $M_l^{i\to j}$$(1\leq i\neq j\leq 3)$ are different from the soft mask. Besides the structure of the proposed MEAN, loss functions are also important for the data-driven approach. It is very important to guarantee the fidelity of the reconstructed image $Z^{i\to j}$ with to the ground truth image $Z_T^{i \to j}$. A reconstruction loss function $L_r$ is
\begin{align}
	L_r=\sum_{i,j}\sum_p \|Z^{i\to j}(p) - Z_T^{i \to j}(p)\|_1.
\end{align}

Color distortion is a possible issue for pixels in the underexposed and overexposed regions of $Z^{i\to j}$ \cite{1lizg2014}.  To remove the color distortion, another loss function $L_c$ is given as
\begin{align}
	L_c=\sum_{i,j}\sum_p\angle (Z^{i\to j}(p), Z_T^{i \to j}(p)),
\end{align}
where  $\angle(Z^{i\to j}(p), Z_T^{i \to j}(p))$ is the angle between $Z^{i\to j}(p)$ and $Z_T^{i \to j}(p)$ \cite{DeepUPE}.

One more feature-wise loss function $L_f$ is computed as
\begin{align}
	L_f=\sum_{i,j}\frac{{\displaystyle \sum_{t=1}^{W_{k,l}}\sum_{s=1}^{H_{k,l}}}(\phi_{k,l}( Z_T^{i \to j} )_{t,s}-\phi_{k,l}( Z^{i\to j})_{t,s})^2}{W_{k,l}H_{k,l}},
\end{align}
where $W_{k,l}$ and $H_{k,l}$ are the dimensions of the respective feature maps, $\phi_{k,l}(\cdot)$ denotes the feature map which is provided by the $l$-th convolution before the $k$-th pooling layer in the VGG network.

Our overall loss function $L$ is calculated as
\begin{equation}
	L= L_r+w_cL_c + w_fL_f,
\end{equation}
where $w_c$ and $w_f$ are two hyper parameters. Since the reconstruction loss function $L_r$ is much more important than the  other two items, $w_c$ and $w_f$ are empirically selected as $0.01$'s, respectively \cite{1zheng2020}.

The numbers of SRRG and EAGB in the proposed MEAN are set as $5$ and $4$, respectively. The MEAN is trained using  an Adam optimizer. The  batch size is 8. The learning rate is initially set to $10^{-5}$, then it is decreased through a cosine annealing method. Our MEAN is trained for 1000 epochs.

Our neural augmentation framework is convergent faster than the corresponding data-driven approach. Note than the data-driven approach only generates a refined term for the physics-driven approach. As such, the amount of training data is reduced by our new framework  to achieve an accuracy  \cite{1nir2021}. Moreover, the physics-driven approach is independent of the training data, our new framework can be applied to to obtain a more general representation of the data than the corresponding data-driven approach.

Once all the images $Z^{i\to j}(1\leq i\neq j\leq 3)$ are generated by the proposed neural augmentation framework, the three panoramic LDR images with different exposures are first produced as in the equations (\ref{fusioneq1})-(\ref{fusioneq2}), and  will then be combined together to generate an information enriched panoramic LDR image in the next section.

\section{Multi-scale Fusion of Panoramic LDR Images with Different Exposures}
\label{localstitching}

The three panoramic LDR images  with different exposures $\tilde{Z}_l(l=1,2,3)$ are merged together as $\tilde{Z}_s$ via the MEF algorithm in \cite{1kou2017} which is on top of edge-preserving smoothing pyramids \cite{21kou2015,2lizg2014}. Since the high frequency information could be lost by both the data-driven approach and the MEF, a log domain physics-driven approach is adopted to preserve the genuine high-frequency information in the fused image for each group of inputting images.

For simplicity, let $\log_2(Z_l+1)$ be denoted as $\hat{Z}_l$. Inputs of the proposed quadratic optimization based enhancement algorithm are the stitched image $\tilde{Z}_s$ and a guidance vector field $V$.  Output is a high-frequency information enhanced image $\hat{Z}$. The vector field $V$ is generated by extending the gradient domain alpha blending as
\begin{align}
	\left\{\begin{array}{ll}
		\nabla \hat{Z}_l(p); & \mbox{if~} p\in \chi_l\\
		\nabla \hat{Z}_{l}(p)+ w_{l-1}^{l}(p)(\nabla \hat{Z}_{l-1}(p)-\nabla \hat{Z}_{l}(p)); &\mbox{if~}p\in \Xi_{l-1}^{l}\\
		\nabla \hat{Z}_{l+1}(p)+ w_{l}^{l+1}(p)(\nabla \hat{Z}_{l}(p)-\nabla \hat{Z}_{l+1}(p)); &\mbox{if~}p\in \Xi_{l}^{l+1}\\
	\end{array}
	\right.,
\end{align}
where the weight $w_{l-1}^{l}(p)$ is calculated as
\begin{align}
	w_{l-1}^{l}(p)=\frac{C_{l-1}^{l}+\theta-x}{2\theta}\; ;\;|x-C_{l-1}^{l}|\leq \theta,
\end{align}
$C_{l-1}^{l}$ is is the center of the overlap area $\Xi_{l-1}^l$. $2\theta$ is the width of $\Xi_{l-1}^{l}$.

The genuine high-frequency information can then be extracted via:
\begin{align}
	\label{high-frequency}
	\min_{\tilde{Z}_d}\{\|\tilde{Z}_d\|_2^2+\lambda(\|\frac{\alpha V_x-\nabla_x \tilde{Z}_d}{\psi(V_x)}\|_2^2+\|\frac{\alpha V_y-\nabla_y \tilde{Z}_d}{\psi(V_y)}\|_2^2)\},
\end{align}
where $\lambda$ and $\alpha$ are empirically chosen as 0.125 and 1.125, respectively, and $\psi(z)$ is
\begin{align}
	\label{psiz}
	\psi(z)=\sqrt{|z|^{0.75}+2}.
\end{align}
The high-frequency information enhanced image $\hat{Z}$ is finally calculated as $\tilde{Z}_s\odot 2^{\tilde{Z}_d}$ with the extracted high-frequency information as $\tilde{Z}_s\odot (2^{\tilde{Z}_d}-1)$. It should be pointed out that the three differently exposed images $\tilde{Z}_l (1\leq l\leq 3)$ can be fused by using an unsupervised learning based multi-scale exposure fusion algorithm. This problem will be studied in our future research.

\section{Experiment Results}
\label{experiment}
Three main contributions of this paper: the proposed WHA, the new neural augmentation framework, and our panoramic HDR stitching algorithms are evaluated in this section.

\textcolor{blue}{The proposed algorithm is implemented using the PyTorch framework on 4 NVIDIA A6000 GPUs. Training data is augmented by mirroring and random cropping of $128 \times 128$ patches from each input image. The networks are trained with the proposed loss functions and an Adam optimizer, using a batch size of 8. The learning rate is initially set to $10^{-5}$ and subsequently decreased via a cosine annealing schedule. The networks are trained for 1000 epochs and evaluated on the validation dataset.}

\subsection{Evaluation of the Proposed WHA}
The accuracy and computational cost of the physics-driven approach are important for our neural augmentation. In this subsection, the proposed WHA is compared with existing IMF estimation algorithms including HHM \cite{1ding2021}, PCRF \cite{CMF-PCRF2018}, TG \cite{CMF-TG2014}, AM \cite{CMF-AM2014}, 3MS \cite{CMF-3MS2013}, GPS \cite{CMF-GPS2017}, MV \cite{CMF-2001}, GC \cite{Zhu}, and  CHM \cite{Grossberg2003} by using the VETHDR-Nikon dataset \cite{1zheng2020}. The PSNR,  SSIM,  FSIM,  iCID and running time are chosen as the evaluation metrics.  All the algorithms are tested by using the Matlab R2019a on a laptop with Intel Core i7-9750H CPU 2. 59GHz and  32. 0 GB memory.

In the VETHDR-Nikon dataset \cite{1zheng2020},  each pair of differently exposed LDR images is well-aligned.  In order to simulate cases with camera movements,  those images in the data-set are cropped according to  the following two equations:
\begin{align}
	\{(x,y)|11\leq x\leq W, 0\leq y\leq H-10\},\\
	\{(x,y)|0\leq x\leq W-10, 11\leq y\leq H\},
\end{align}
where $W$ and $H$ are the weight and height of the images in the dataset. As demonstrated in Table \ref{tab:exp_IMFs}, the proposed WHA achieves the best performance from all the PSNR,  SSIM,  FSIM,  and iCID points of view, and its speed is slightly slower than the  CHM \cite{Grossberg2003}.

\begin{table}[htbp]
	\setlength{\abovecaptionskip}{0pt}
	\setlength{\belowcaptionskip}{0pt}
	\centering
	\scriptsize
	\caption{Comparison of different intensity mapping algorithms. The best results and the second-based ones are shown in bold, respectively
		\label{tab:exp_IMFs}
	}\rowcolors[]{1}{white!20}{gray!15}
	\resizebox{1\columnwidth}{!}
	{
		\begin{tabular}{c|ccccc}
			\toprule[1pt]
			\textbf{Method} & \textbf{PSNR$\uparrow$} & \textbf{SSIM$\uparrow$} & \textbf{FSIM$\uparrow$} & \textbf{iCID(\%)$\downarrow$} & \textbf{Time(s)$\downarrow$}  \\
			\midrule
			HHM \cite{1ding2021} & 29.88 & 0.8977 & 0.9673 & 8.24  & 0.11  \\
			PCRF \cite{CMF-PCRF2018} & 30.69 & {\color[HTML]{FF0000} 0.9056} & 0.9715 & 7.67  & 1.79  \\
			TG \cite{CMF-TG2014} & 29.34 & 0.8897 & 0.9391 & 11.35  & 12.52  \\
			AM \cite{CMF-AM2014}& 25.68 & 0.8413 & 0.8882 & 18.49  & 0.54  \\
			3MS \cite{CMF-3MS2013}& 28.96 & 0.8857 & 0.9455 & 11.17  & 7.70  \\
			GPS \cite{CMF-GPS2017} & 30.96 & 0.8945 & 0.9643 & 9.36  & 151.09  \\
			MV \cite{CMF-2001} & 28.14  & 0.8964  & 0.9470  & 10.28  &  0.66  \\
			GC \cite{Zhu}& 28.17 & 0.8693 & 0.9385 & 11.98  & 1.77  \\
			CHM \cite{Grossberg2003} & {\color[HTML]{FF0000} 32.36} & 0.9029 & {\color[HTML]{FF0000} 0.9740} & {\color[HTML]{FF0000} 7.10} & \textbf{0.08}  \\
			WHA & \textbf{34.38} & \textbf{0.9153} & \textbf{0.9815} & \textbf{4.77}  &  {\color[HTML]{FF0000} 0.08}  \\
			\bottomrule[1pt]
		\end{tabular}
	}
\end{table}

\begin{figure*}[htbp]
	\setlength{\abovecaptionskip}{0pt}
	\setlength{\belowcaptionskip}{0pt}
	\begin{center}
		\includegraphics[width=1.0\linewidth]{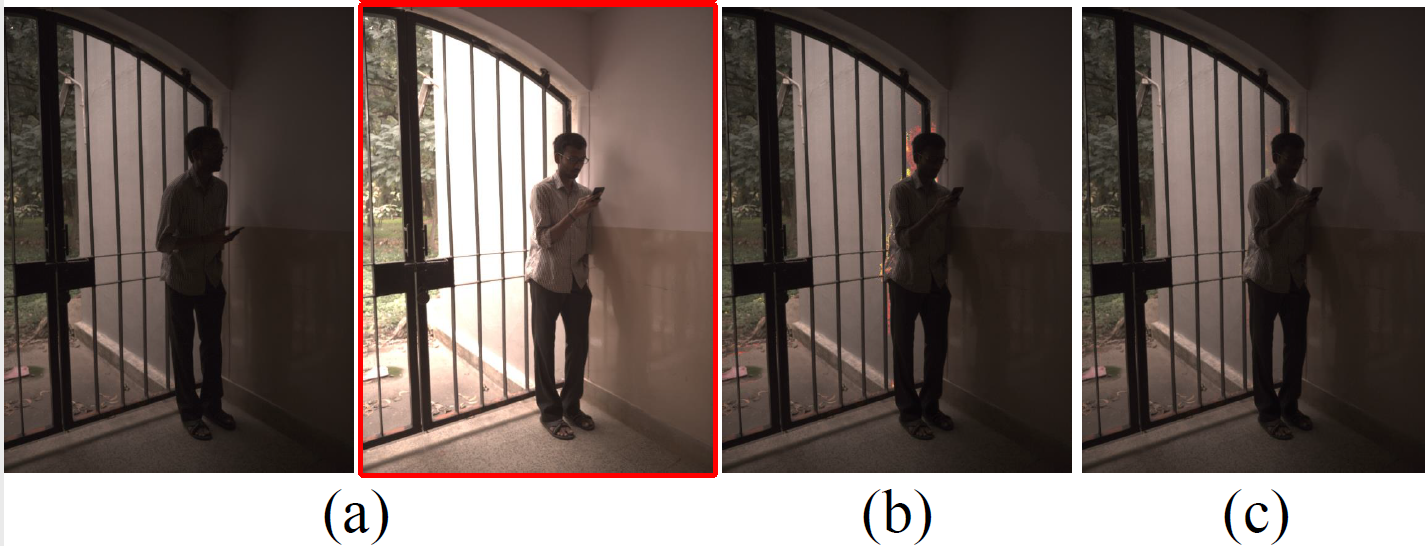}
	\end{center}
	\caption{Comparison of the impact of CHM and WHA on the ghost removal algorithm in \cite{1lizg2014}. (a) One pair of differently exposed LDR images with a moving human subject from the dataset \cite{deghost_dataset2}, (b) the result of the ghost removal algorithm in \cite{1lizg2014} by using the CHM in \cite{Grossberg2003}, (c) the result of the ghost removal algorithm in \cite{1lizg2014} by using the proposed WHA.  The image in the red boxes is the selected reference images}
	\label{fig:exp_deghost}
\end{figure*}

The proposed WHA can be applied to investigate ghost removal of differently exposed LDR images. There are many ghost removal algorithms for differently exposed images \cite{1lizg2014}. The proposed WHA is adopted to improve the ghost removal algorithm in \cite{1lizg2014}. One of the input images is selected as the reference image. Pixels in all other images are divided into inconsistent and consistent pixels. All the inconsistent pixels are corrected by using the correlation between the reference image and the corresponding image. The IMFs which are estimated by the CHM in \cite{Grossberg2003} are used in \cite{1lizg2014} to detect all the inconsistent pixels and correct them. The CHM in \cite{Grossberg2003} is replaced by the proposed WHA. As shown in Fig. \ref{fig:exp_deghost}(b), there is serious color distortion in the wall by using the CHM in \cite{Grossberg2003}.  Fortunately, the color distortion is significantly reduced by our WHA. It should be pointed out that the WHA was recently applied in \cite{1liuz2023} to design an optical flow estimation algorithm for differently exposed LDR images in LDR domain. It was shown in \cite{1liuz2023} that the WHA outperforms the gamma correction which was widely utilized by deep-learning based ghost removal algorithms.

\subsection{Ablation Study}
Four important components of the proposed neural augmentation frameworks are the physics-driven approach, the EAGB, the short-skip connection, and the multi-scale refinement. Their performances are evaluated in this subsection.
\begin{table*}[htbp]
	\setlength{\abovecaptionskip}{0pt}
	\setlength{\belowcaptionskip}{0pt}
	\centering
	\scriptsize
	\caption{\textcolor{blue}{Ablation study of the data-driven approach on generating high exposure images from low exposure ones ($\uparrow$: larger is better)}
		\label{tab3}
	}\rowcolors[]{0.1}{white!20}{gray!15}
	\resizebox{1.\columnwidth}{!}
	{
		\begin{tabular}{c|cccc|ccc}
			\toprule[0.8pt]
			\textbf{Case} & \textbf{physics-driven } & \textbf{EAGB} & \textbf{short-skip} & \textbf{ multi-scale } & \textbf{SSIM ($\uparrow$)} & \textbf{PSNR (dB: $\uparrow$)}  & \textbf{FSIM ($\uparrow$)} \\
			\midrule
			1 & Y & N & N & N  & 0.9361  & 31.28  & 0.9800\\
			2 & N & N & Y & Y  & 0.9420  & 31.86  & 0.9804 \\
			3 & N & Y & Y & Y  & 0.9428  & 32.18  & 0.9786 \\
			4 & Y & N & Y & Y  & 0.9434  & 32.30  & 0.9811\\
			5 & Y & Y & N & Y  & 0.9448  & 33.09  & 0.9816\\
			6 & Y & Y & Y & N  & 0.9461  & 33.27  & 0.9817\\
			7 & Y & Y & Y & Y  & \textbf{0.9464}  & \textbf{33.39}  & \textbf{0.9822}\\
			\bottomrule[1pt]
		\end{tabular}
	}
\end{table*}

{\it With vs without the physics-driven approach}: Our neural augmentation framework and the data-driven approach are compared with each other. The data-driven approach is almost identical to the data-driven approach of our neural augmentation framework, except for disabling the EAGB. This is because that the initial synthetic images is required by the EAGB. Table \ref{tab3} demonstrates that our framework can enhance the SSIM, PSNR, and FSIM. Additionally, Fig. \ref{Figure_7} illustrates that our new framework converges faster than the data-driven approach. It is noteworthy that case 1 in Table \ref{tab3} is the physics-driven approach. Clearly, the data-driven approach can significantly enhance the physics-driven approach.

{\it With vs without the EAGB}:  As demonstrated in Fig. \ref{Figure_7},  the EAGB can be applied to improve the stability of the proposed framework, and it can also achieve higher PSNR and SSIM values at different epochs.

{\it With vs without the short-skip-connection}: As shown in Fig. \ref{Fig_4}, our proposed SRRG can transmit information from shallow layers to deep layers. The PSNR and SSIM at different epochs are depicted in Fig. \ref{Figure_7}, indicating that the mode with the short-skip connection can obtain superior results.

{\it Multi-scale vs single-scale}: The proposed multi-scale framework and another mode with a single-scale structure are compared with each other. The multi-scale structure achieves higher SSIM, PSNR, and FSIM values.

\subsection{Evaluation of the Depth of Network}
\textcolor{blue}{The physical model enhances the interpretability of the entire algorithm. By focusing on learning the residuals, the network complexity can be reduced. This section presents experiments on the network depth. The results, shown in Table \ref{Depth}, indicate that with too few layers, the network's representational capacity is insufficient, while with too many layers, additional training epochs are required.}

\label{Depth}
\begin{table*}[htbp]
	\setlength{\abovecaptionskip}{0pt}
	\setlength{\belowcaptionskip}{0pt}
	\centering
	\scriptsize
	\caption{\textcolor{blue}{Ablation study of the Depth of Network  ($\uparrow$: larger is better)}
		\label{Dep}
	}\rowcolors[]{0.1}{white!20}{gray!15}
	\resizebox{0.7\columnwidth}{!}
	{
		\begin{tabular}{c|ccc}
			\toprule[0.1pt]
			\textbf{Number of SRRG}  & \textbf{SSIM ($\uparrow$)} & \textbf{PSNR (dB: $\uparrow$)}  & \textbf{FSIM ($\uparrow$)} \\
			\midrule
			2   & 0.9434  & 32.92  & 0.9811 \\
			4   & 0.9464  & 33.39  & 0.9822 \\
			6   & 0.9453  & 33.25  & 0.9815\\
			\bottomrule[0.1pt]
		\end{tabular}
	}
\end{table*}

\subsection{Evaluation of Different Physics-driven Approaches}
\label{IMFaccuracy}
It was shown in \cite{1nir2021} that the physics-driven approach can be improved by the data-driven approach in the neural augmentation. However, one important issue was ignore in  \cite{1nir2021}, i.e., is the accuracy of the physics-driven approach important for the neural augmentation? Experimental results are given to address the issue in this subsection. More specifically, highly accurate IMFs are compared with lowly accurate IMFs.
\begin{figure*}[htb]
	\centering
	\includegraphics[width=1\textwidth]{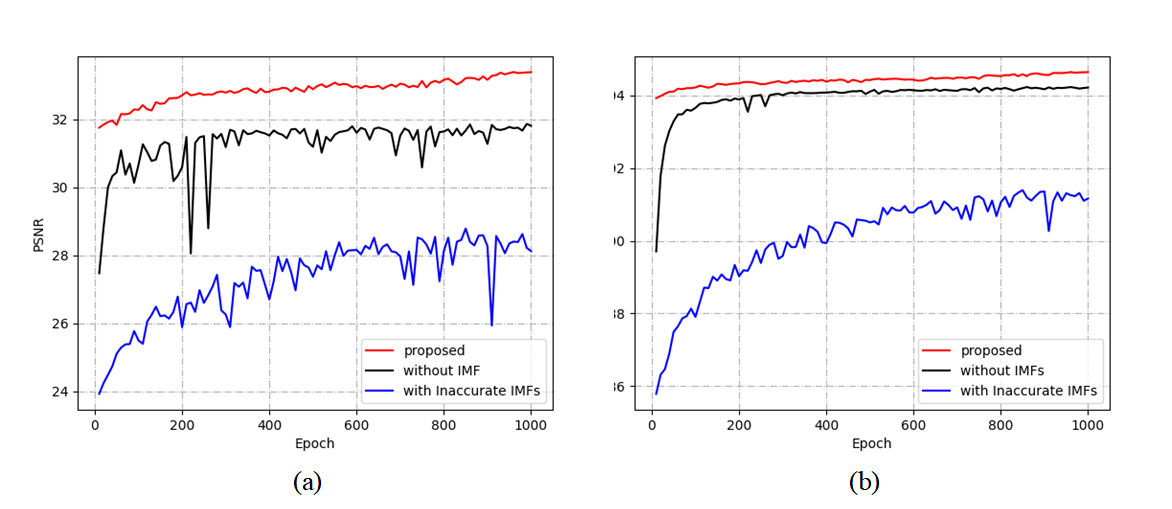}
	\caption{Comparison of two different neural augmentation framework and the data-driven approach. The accuracy of the physics-driven approach is indeed very important for the proposed neural augmentation framework.}
	\label{Figure_7}
\end{figure*}
Two different overlapping area, denoted as R1 ($480\times 200$) and R2 ($480\times 100$), are compared to answer the question. The IMFs of the former are more accurate than those of the latter. As depicted in Fig. \ref{Figure_7}, the former converges faster than the latter. The PSNR and SSIM of R1 are also increased. Therefore, the accuracy of the physics-driven approach is important for the proposed neural augmentation framework. To reduce the inference complexity of the neural augmentation, the physics-driven approach is also required to be simple.

It is also demonstrated in Fig. \ref{Figure_7} that data-driven approach is typically locally stable. Thus, a neural augmentation on top of an accurate physics-driven approach usually outperforms the corresponding data-driven approach.

\begin{table}[htbp]
	\setlength{\abovecaptionskip}{0pt}
	\setlength{\belowcaptionskip}{0pt}
	\centering
	\scriptsize
	\caption{\textcolor{blue}{MEF-SSIM of seven different stitching algorithms}
		\label{tabfusion}
	}\rowcolors[]{2}{white!20}{gray!15}
	\resizebox{0.9\columnwidth}{!}
	{
		\begin{tabular}{c|ccccccc}
			\toprule[1.2pt]
			\textbf{Set} & \textbf{ \cite{Brown} } & \textbf{\cite{1liao2020}} & \textbf{\cite{1ding2021}} & \textbf{\cite{Jia}}  & \textbf{\cite{Lukas2022}} & \textbf{Ours}  & \textbf{Ours with (\ref{high-frequency})}\\
			\midrule
			\multirow{1}*{1}    & 0.4543   & 0.8036         & 0.9380   & 0.5436   & 0.5257      &0.9723   &\textbf{0.9726}   \\
			\multirow{1}*{2}    & 0.4552   & 0.7048         & 0.8948   & 0.6850   & 0.8621      &0.9904   &\textbf{0.9905} \\
			\multirow{1}*{3}    & 0.5904   & 0.6918         & 0.9467   & 0.4614   & 0.6811      &0.9941   &\textbf{0.9943} \\
			\multirow{1}*{4}    & 0.7034   & 0.7386         & 0.9861   & 0.7124   & 0.6944      &0.9902   &\textbf{0.9904} \\
			\multirow{1}*{5}    & 0.1567   & 0.4391         & 0.8239   & 0.2877   & 0.5218      &0.8499   &\textbf{0.8532} \\
			\multirow{1}*{6}    & 0.5829   & 0.5843         & 0.9336   & 0.6058   & 0.7264      &0.9666   &\textbf{0.9678} \\
			\multirow{1}*{7}    & 0.2314   & 0.4086         & 0.9537   & 0.6549   & 0.9392      &0.9929   &\textbf{0.9936} \\
			\multirow{1}*{8}    & 0.3080   & 0.2690         & 0.9413   & 0.3029   & 0.7266      &0.9186   &\textbf{0.9215} \\
			\multirow{1}*{9}    & 0.4198   & 0.7182         & 0.9663   & 0.5743   & 0.7511      &0.9920   &\textbf{0.9923} \\
			\multirow{1}*{10}   & 0.1778   & 0.5477         & 0.9417   & 0.4098   & 0.6686      &0.9844   &\textbf{0.9849} \\
			\multirow{1}*{11}   & 0.4759   & 0.6925         & 0.9573   & 0.7375   & 0.7978      &0.9906   &\textbf{0.9908} \\
			\multirow{1}*{12}   & 0.3331   & 0.3390         & 0.8885   & 0.3083   & 0.6654      &0.9424   &\textbf{0.9425} \\
			\multirow{1}*{13}   & 0.4064   & 0.5268         & 0.9758   & 0.4483   & 0.7771      &0.9854   &\textbf{0.9857} \\
			\multirow{1}*{14}   & 0.2590   & 0.6723         & 0.9516   & 0.3132   & 0.7055      &0.9778   &\textbf{0.9783} \\
			\multirow{1}*{15}   & 0.3470   & 0.5467         & 0.9503   & 0.4174   & 0.6720      &0.9909   &\textbf{0.9914} \\
			\hline
			\multirow{1}*{avg}  & 0.3829   & 0.5789         & 0.9366   & 0.4975    &0.7143     &0.9692   &\textbf{0.9700}\\
			\bottomrule[1pt]
		\end{tabular}
	}
\end{table}
\subsection{Comparison of Different Panoramic Stitching Algorithms}
The proposed panoramic HDR stitching algorithm is compared with four state-of-the-art (SOTA) panoramic stitching algorithms in \cite{Brown,1liao2020,1ding2021,Jia, Lukas2022}.

\begin{figure*}[htb]
	\centering
	\includegraphics[width=1\textwidth]{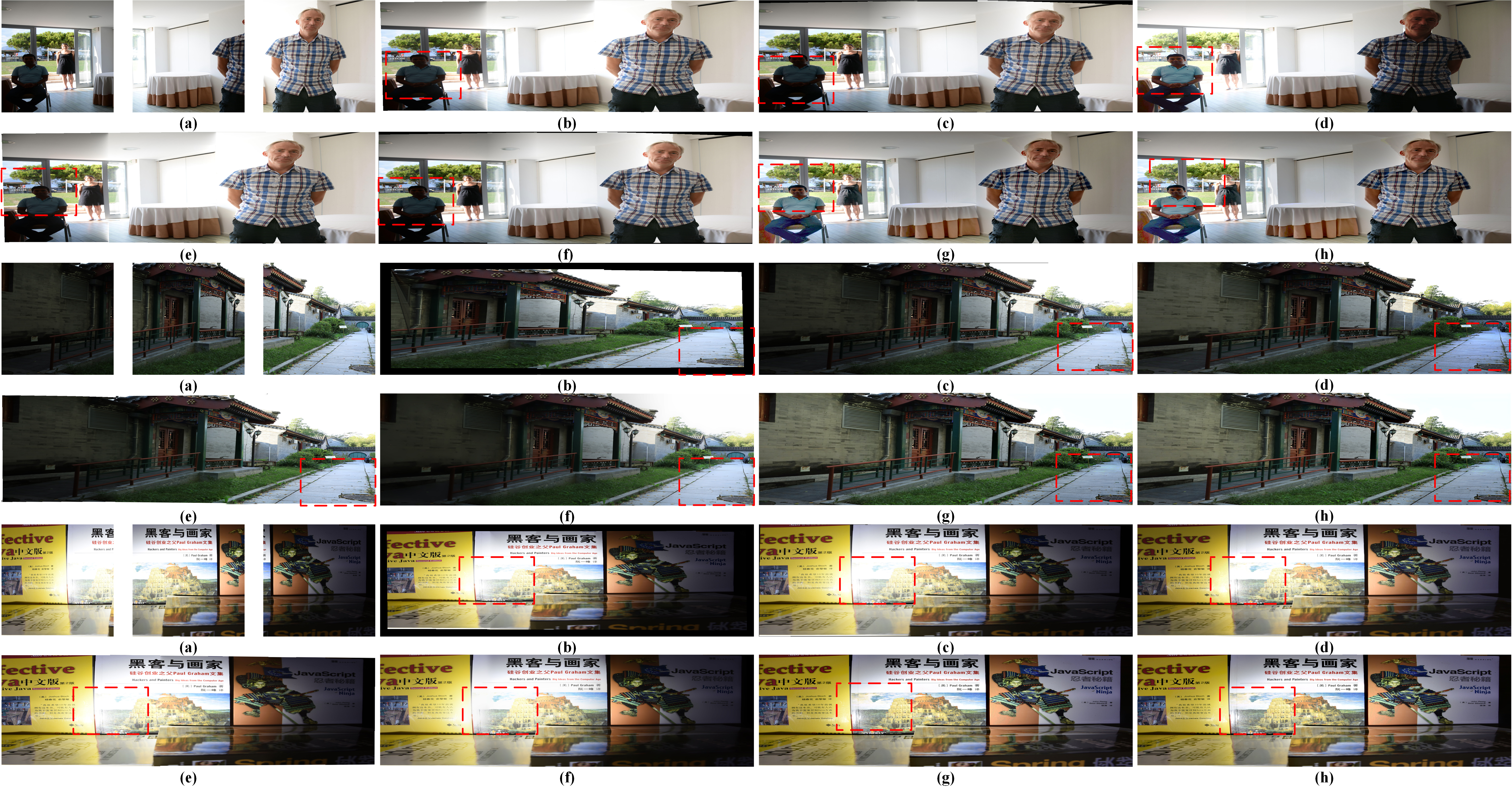}
	\caption{ \textcolor{blue}{Comparison of different panoramic stitching algorithms. (a) are three inputting images with different exposures and pairwise OFOVs; Panoramic LDR images produced by (b) \cite{Brown}, (c) \cite{1liao2020}, (d) \cite{1ding2021}, (e) \cite{Jia}, (f) \cite{Lukas2022}, (g) our algorithm without Eq. (\ref{high-frequency}), and (h) our algorithm with Eq. (\ref{high-frequency}).} }
	\label{Fusion}
\end{figure*}

As illustrated in Fig. \ref{Fusion}, serious brightness inconsistencies are evident in the stitched images produced by the algorithms in \cite{Brown,1liao2020,Jia, Lukas2022}. These inconsistencies are overcome by the algorithms in \cite{1ding2021} and our algorithm. Information in the darkest regions of the first set and information in the brightest regions of the second set are preserved better using our algorithm compared to the algorithms in \cite{Brown,1liao2020,Jia,1ding2021, Lukas2022}. Additionally, as highlighted by the red boxes, the algorithm in \cite{1ding2021} produces color distortions. The MEF-SSIM metric in \cite{1ma2015} is also employed to compare the all these. The MEF-SSIM is computed by using the ground-truth images of three panoramic LDR images with different exposures and a panoramic LDR image stitched by one of these algorithms. As shown in Table \ref{tabfusion}, our algorithm noticeably outperforms those in \cite{Brown,1liao2020,1ding2021,Jia, Lukas2022} from the MEF-SSIM perspective. Our algorithm (\ref{high-frequency}) extract real high-frequency information from each set of images and can further enlarge the average MEF-SSIM. Clearly, the HDR stitching is highly demanded for real-world HDR scenes.

\subsection{Evaluation of the Local Stitching Algorithm (\ref{high-frequency})}
\begin{figure*}[htb]
	\centering
	\includegraphics[width=1.0\textwidth]{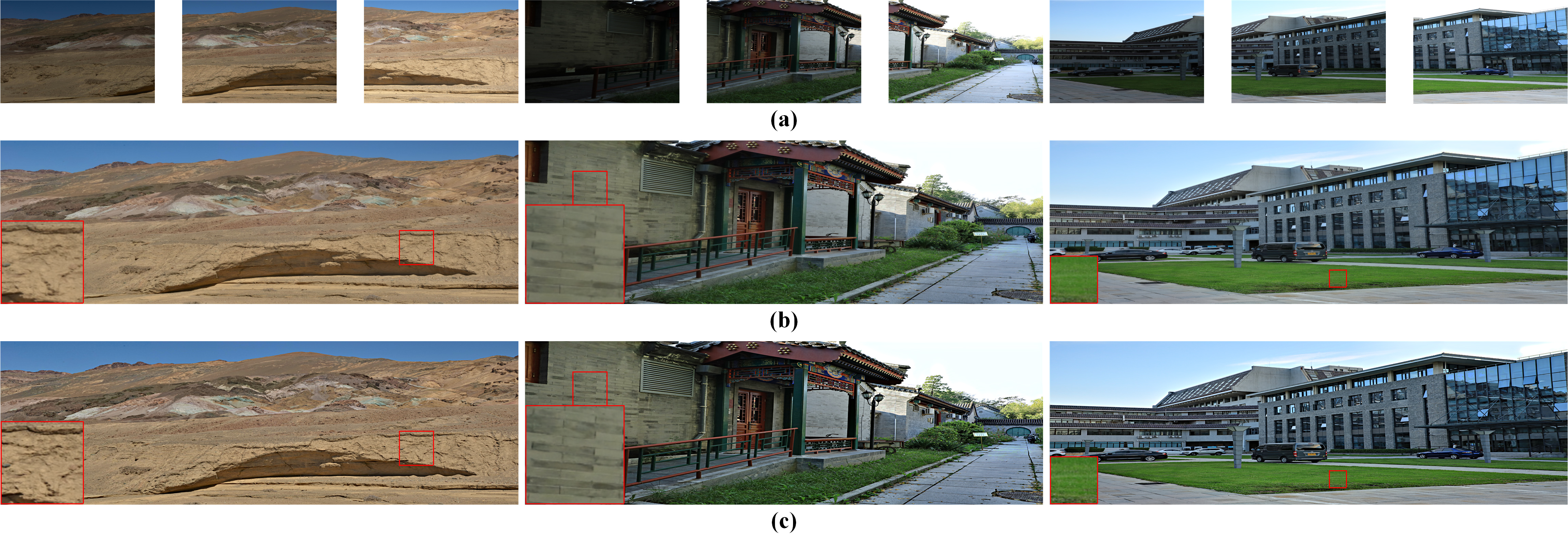}
	\caption{Evaluation of the proposed local stitching algorithm. Real high-frequency information is extracted by the physics-optimization based algorithm (\ref{high-frequency}). Thus, the resultant images become sharper, and MEF-SSIM values are also enlarged. }
	\label{Fusion_Detail}
\end{figure*}
The proposed local stitching algorithm (\ref{high-frequency}) is evaluated in this subsection. As shown by enlarged parts in Fig. \ref{Fusion_Detail}, high-frequency components such as sharp edges, textures, fine details, etc are indeed preserved better by the local one. Therefore, the local algorithm can indeed be applied to achieve {\it instance adaptation}. It should be mentioned that different users have different preferences on the sharpness of the enhanced image $\hat{Z}$. This can be achieved via a weighted combination $(\nu \tilde{Z}_s+(1-\nu)\hat{Z})$ with $\nu$ in the interval $[0, 1]$. An interactive mode can be provided by allowing tuning the $\mu$ on-line.

\subsection{A New Panoramic HDR Stitching System}
\label{new}
\begin{figure*}[htb]
	\centering
	\includegraphics[width=1.0\textwidth]{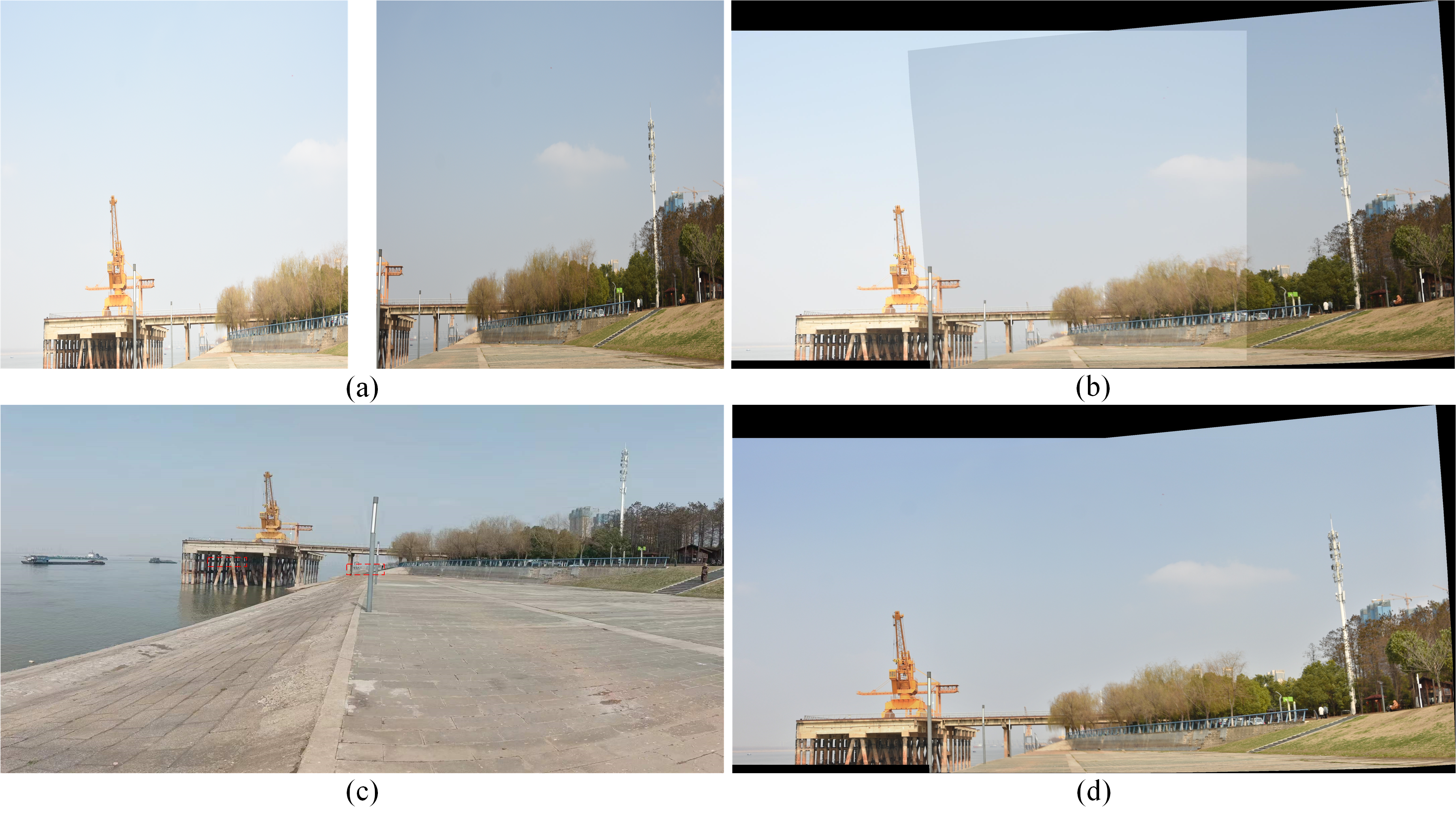}
	\caption{Comparison of different panoramic stitching systems. (a) are two inputting images captured with different exposures and pairwise OFOVs;  (b) is the Panoramic LDR image produced by \cite{nie2023parallax}; (c) is the Panoramic LDR image taken with a smart phone; (d) is the Panoramic LDR image produced by our new  panoramic HDR stitching method.}
	\label{Fig_new}
\end{figure*}
A new panoramic stitching system is proposed in this subsection. Instead of capturing a series of images by existing panoramic stitching systems, multiple differently exposed images are captured with varying orientations. For simplicity, the case of two differently exposed images is studied in this subsection. The new system can be easily extended to the case of three different exposed images. The differences between the orientations of these two images are about 15 degrees as shown in Fig. \ref{Fig_new}. Since the exposures of the two images are different, the algorithm in \cite{nie2023parallax} cannot be applied to align these images directly. The right images is first mapped to the brightness of the left one by using the IMFs between them. They are then aligned by simultaneously using homography transformations and thin-plate spline transformations. The homography transformation provides a global linear transformation, while the thin-plate spline transformation allows local nonlinear deformation. These two images are finally stitched by using the proposed neural augmentation based algorithm. Two panoramic LDR images with different exposures are produced by using the same set of homography transformation and thin-plate spline transformations. They are fused to generate the final image as shown in Fig. \ref{Fig_new} (d).

Such a new panoramic HDR stitching system is compared with two conventional panoramic LDR stitching systems. One is from a smart phone and the other is the algorithm in \cite{nie2023parallax}. The exposure times of those images are supposed to be almost the same for the conventional panoramic LDR stitching systems. As demonstrated in Fig. \ref{Fig_new}, the images produced by the new panoramic HDR stitching system is much better than those generated by the conventional panoramic LDR stitching systems. The panoramic LDR image captured with a mobile phone exhibits noticeable distortion in the bridge deck, as highlighted by these two red boxes. The panoramic LDR image generated using \cite{nie2023parallax} includes visible seams, and has three different brightnesses in different regions. These artifacts are removed by using the proposed panoramic HDR stitching algorithm.

\section{Conclusion Remarks and Discussions}
\label{conclusion}
High dynamic range imaging and panoramic stitching are two popular types of computational photography. They are studied jointly in this paper while they are usually investigated independently. There are many applications of high dynamic range (HDR) panoramic imaging. A novel algorithm is proposed in this paper to stitch multiple geometrically synchronized and differently exposed sRGB images with overlapping fields of views (OFOVs) for a real-world high dynamic range (HDR) scene. Different exposed images of each view are initially generated by using an accurate  physics-driven approach, are then refined by a data-driven approach, and are finally adopted to generate panoramic LDR images with different exposures. All the panoramic LDR images with different exposures are  merged using a multi-scale exposure fusion algorithm, incorporating a physics-driven approach for the enhancement of high-frequency information, resulting in an information-enriched panoramic LDR image.

It should be pointed out that the inputs are required to have sufficiently large overlapping fields of views large OFOVs by the proposed neural augmentation. It would be interesting to consider the case that the OFOVs are not large. This challenging case will be investigated by exploring new ways to derive the highly accurate physics-driven approach required by the neural augmentation. One more interesting problem is to consider the case that inputs are three differently exposed raw images with OFOVs from an HDR scene. Our proposed framework would be extended to study this problem in future.

\end{document}